# Природа явления «на кончике языка»: неадекватная локализация нейронной сети или разрыв ее межнейронных связей ?

## Гопыч П.М.

### Харьковский национальный университет им. В.Н. Каразина
### пл. Свободы 4, Харьков 61077 Украина, pmg@kharkov.com

На основе недавно предложенной трехэтапной количественной нейросетевой модели исследована возможность возникновения состояний «на кончике языка» (НКЯ) по причине разрыва части межнейронных связей активной сети. На численном примере показано, что такие состояния возникают с вероятностью в $(1.5 \pm 0.3) \cdot 10^3$ раз меньшей, чем вероятность их возникновения из-за неадекватной (неполной) локализации сети. Показано, что разрыв межнейронных связей сети не может быть причиной возникновения затяжных состояний НКЯ.

On the base of recently proposed three-stage quantitative neural network model of the tip-of-the-tongue (TOT) phenomenon a possibility to occur of TOT states coursed by neural network links' disruption has been studied. Using a numerical example it was found that TOTs coursed by interneron links' disruption are in $(1.5 \pm 0.3) \cdot 10^3$ times less probable then those coursed by irrelevant (incomplete) neural network localization. It was shown that delayed TOT states' etiology cannot be related to neural network interneuron links' disruption.

## 1. Введение

Широко известное в психолингвистике явление «на кончике языка» (НКЯ) проявляется как свойство памяти: в состоянии НКЯ человек ощущает, что знает вспоминаемое слово и уверен, что может его вспомнить, но именно в данный момент оно ему временно недоступно. Таким образом, НКЯ проявляется как дефект словарной (вербальной) памяти человека. В настоящее время при анализе речевого воспроизведения слов сосуществуют две исследовательские традиции [6]: основанная на исследовании речевых ошибок и на хронометрии наименования изображений. Модели явления НКЯ возникли из традиции по исследованию речевых ошибок и развиваются в рамках трех исследовательских подходов [7]: психолингвистического, метакогнитивного и основанного на исследованиях памяти.

Обе указанные традиции и все три упомянутые подходы объединяет трехэтапная количественная нейросетевая модель явления НКЯ [2,4], предложенная на основе нейросетевой модели памяти [1], существенно использующей теорию проверки статистических гипотез. Модель может количественно объяснить многие эффекты, присущие НКЯ: семантическое предпочтение; существование негативных, иллюзорных и затяжных состояний НКЯ; вспоминание частичной информации об искомом слове (например, первой буквы или рода); зависимость состояний НКЯ от возраста человека; состояния НКЯ у больных с повреждениями мозга и т.д. Она свидетельствует также в поддержку обоснованности сомнений Тулвинга [8] в справедливости «доктрины о согласии». Кроме того, модель позволяет явно описать [5] некоторые функции сознания (например, возникновение «чувства знания» [5]), а также проявления элементов неявно-автоматического (с относительно низкой степенью осознания) и осознанного (контролируемого волевым образом при посредничестве осознания) поведения.

Ранее [2] в качестве причины возникновения состояний НКЯ рассматривалась только возможность неполной (неадекватной) локализации рабочей нейронной сети из состава активного словарного узла. В настоящей работе методом компьютерного моделирования исследована возможность появления таких состояний, вызванных разрывами части межнейронных связей рабочей сети, и круг опытов, в которых такие НКЯ могут себя проявлять.

## 2. Краткое описание модели

Существующие компьютерные модели воспроизведения речи являются нейросетевыми моделями, узлы которых представляют собой целостные лингвистические единицы [6]. Согласно нашей модели для выбранного словарного узла его нейронная сеть строится из нескольких взаимосвязанных обученных двухслойных автоассоциативных искусственных нейронных сетей (ИНС), каждая из которых предназначена для представления отдельных семантических, лексических или фонологических компонент слова (такое многомерное представление элементов словарной памяти согласуется с имеющимися данными о распределении активности человеческого мозга, проявляемой им при обработке речи [9]). Эти ИНС оперируют с образами сигналов, кодирующими конкретную лингвистическую информацию, и представляют собой наборы положительных и отрицательных единиц, которые моделируют нервные импульсы воздействующие на возбудительные (+1) и тормозные (−1) синапсы живых нейронов [1].

Предлагаемая модель предполагает, что формирование и разрешение состояний НКЯ проходит три следующих этапа [2,4]:

1. *Локализация словарного узла,* когда на основе предоставляемой семантической информации выделяется дерево упомянутых выше обученных ИНС (это взаимосвязанные элементы памяти), представляющих информацию, относящуюся к вспоминаемому слову. Поскольку состояния НКЯ возникают обычно при определении трудных или редко встречающихся слов предполагаем, что соответствующие обученные ИНС (хранящаяся в них информация) могут быть повреждены (например, вследствие недостаточной консолидации памяти) и/или неправильно выделены (например, из-за недостаточности или неадекватности входной семантической информации). Степень повреждения и/или неполноты локализованной ИНС (запомненной информации) и определяет главным образом специфику возникающего состояния НКЯ.

2. *Извлечение слова (компоненты слова) из памяти.* Процесс извлечения из памяти при свободном вспоминании инициирует серия из случайных наборов положительных и отрицательных единиц (случайных наборов нервных импульсов), поступающих на вход обученной ИНС. Процесс извлечения из памяти при вспоминании с подсказкой такой же, но наборы нервных импульсов, поступающих на вход той же ИНС, уже не являются полностью случайными и содержат фиксированную часть истинной информации о вспоминаемом образе (это подсказка). Результатом каждой попытки извлечения информации из памяти является набор из положительных и отрицательных единиц (набор нервных импульсов) возникающих на выходе ИНС [1].

Конечный этап вспоминания - это *сравнение* образа, возникшего на выходе ИНС, с эталонным образом из метапамяти *и принятие решения* о прекращении или продолжении процесса вспоминания. Если только что всплывший при тестировании обученной сети и эталонный образы совпадают, то вспоминание завершено. В противном случае этап 2 (см. выше) повторяется, другой случайный (или отчасти случайный) набор нервных импульсов поступает на вход той же выделенной ИНС и т.д. пока эталонный образ не будет идентифицирован или процесс вспоминания не будет остановлен по независимым внешним причинам. *Сравнение и принятие решения* являются метакогнитивными процессами (процессами метауровня) и отделены от процесса собственно извлечения информации из памяти (когнитивного процесса предметного уровня).

В [2] было показано, что описанные выше ИНС с частично утраченными входными нейронами могут моделировать обычные и затяжные состояния НКЯ. На примере таких сетей были даны также определения силы и вероятности возникновения НКЯ [2]. Однако возможность возникновения НКЯ в сетях без утраченных нейронов, но с разрывом части межнейронных связей, ранее не рассматривалась.

## 3. Численный пример

Будем анализировать возможность возникновения НКЯ в сети с разрывом части межнейронных связей на примере простейшей двухслойной автоассоциативной ИНС с одинаковым числом $N$ нейронов во входном (выходном) слое и межнейронными связями типа «все со всеми». В качестве составляющих сеть нейронов выбираем традиционные нейроны МакКалоха-Питтса со ступенчатой ответной функцией и нулевым порогом срабатывания, вид эталонного, входных и выходных векторов, обслуживающих сеть, элементы синаптической матрицы и правила преобразования сетью входных сигналов заимствуем из [1].

На примере такой сети с $N = 9$ по результатам точных вычислений было найдено [2], что при утрате ею $N_k = 4$ входных нейронов вероятность возникновения состояния НКЯ составляет 4.8%. Сейчас исследуем возможность появления того же состояния НКЯ в такой же сети без погибших входных нейронов, но с частью разорванных межнейронных связей. Полагаем, что число $N_d$ случайно выбранных разорванных межнейронных связей из их общего числа $N$ х $N = 81$ есть $N_d = 10$. Для сети с заданными $N$ и $N_d$ число разных вариантов возможного выбора разрываемых связей велико: $C^{10}_{81} \approx 2.3 \cdot 10^{11}$. По этой причине вероятность возникновения состояний НКЯ в такой сети будем исследовать приближенно методом многократных статистических испытаний.

Пусть разрываемые межнейронные связи между $i$-м входным и $j$-м выходным нейронами (им соответствуют элементы синаптической матрицы $w_{ij} = 0$) распределены в исследуемой сети случайно и равновероятно. Будем моделировать сети с разными случайными наборами $N_d = 10$ разрываемых связей и для каждой такой сети по методике, описанной в [1], вычислять значение вероятности свободного вспоминания $P = P_{FR}$ запомненного в ней вектора-эталона (в этом случае, см. рис.1а, степень его искажения $d = 1$). По результатам $10^6$ таких испытаний (десять серий по $10^5$ испытаний в каждой) было найдено 66 разных значений $P_{FR}$ и среди них – значение $P_{FR} = 0.28516$, характерное для исследуемого состояния НКЯ, обусловленного утратой $N_k = 4$ входных нейронов той же сети (кривая 3 на рис.1b работы [2]). Одновременно для всех 66 разных значений $P_{FR}$ были оценены и частоты $D(P_{FR})$ их появления (см. гистограмму на рис. 1b). Далее для всех найденных распределений разорванных связей, для которых $P_{FR} = 0.28516$, были вычислены значения вероятности вспоминания $P(d)$ для всех $d$. Оказалось, что они распадаются на две группы (их значения показаны на рис.1а разными крестиками), для одной из которых (кривая 4 на рис. 1а) все значения $P(d)$ совпадают со значениями $P(d)$ для состояния НКЯ, исследованного ранее в [2]. Таким образом, для рассмотренной сети без погибших нейронов, но с $N_d = 10$ разорванными межнейронными связями реализуется состояние НКЯ, характеристики вспоминания которого полностью совпадают с характеристиками вспоминания НКЯ для сети с $N_k = 4$ погибшими входными нейронами, но без разорванных связей.

Из анализа результатов описанного компьютерного моделирования было найдено и соотношение между частотами (вероятностями) реализации значений $P(d)$, ложащихся на рис.1а на кривые 3 и 4. В итоге находим приближенную оценку вероятности появления обнаруженного состояния НКЯ (кривая 4 на рис.1а) в сетях со случайно разорванными межнейронными связями: $(7.0 \pm 1.3) \cdot 10^{-3}$ %. В сетях с частично утраченными входными нейронами аналогичное состояние НКЯ возникает (см. выше) с вероятностью 4.8% (случай $N_k = 4$ утраченных входных нейронов). Следовательно, в сетях с поврежденным слоем входных нейронов НКЯ могут возникать в $(1.5 \pm 0.3) \cdot 10^3$ раз чаще, чем в сетях со случайными разрывами межнейронных связей.

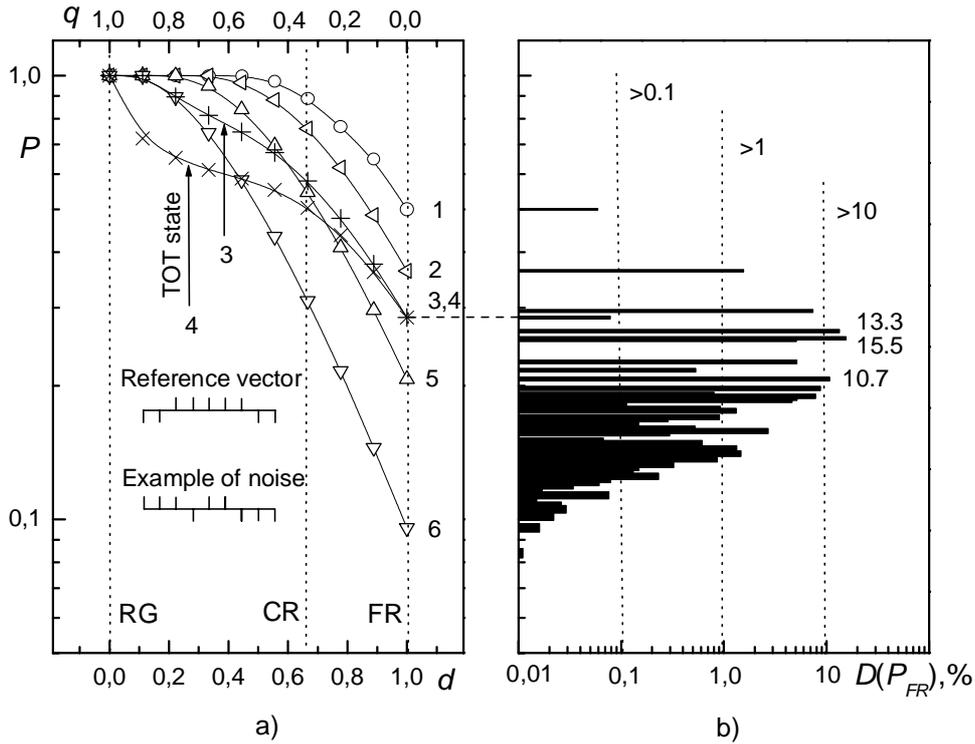

Рис.1a. Вероятность $P(d) = P(1 - q)$ вспоминания запомненного эталона сетью с $N = 9$. Шкала степени искажения эталона $d$ внизу, интенсивности подсказки $q$ вверху. Интерполяционные кривые соединяют точные значения $P$ (разные значки), вычисленные [1] для сети без повреждений (кривая 1) и для той же сети, но с $N_d = 10$ случайно выбранными разорванными межнейронными связями (кривые 2-6). Для кривых 3,4 вероятности $P$ ($d = 1$) = $P_{FR}$ свободного вспоминания одинаковы. Состоянию НКЯ соответствует кривая 4. Все значения $P$ на этой кривой полностью совпадают с соответствующими значениями $P$ для состояния НКЯ, обусловленного утратой в рабочей сети $N_k = 4$ случайно отобранных входных нейронов со всеми своими связями [2]. На вставке показаны эталонный (Reference vector) и пример шумового (Example of noise) векторов (черточки выше горизонтальной линии обозначают их проекции +1, ниже – их проекции –1). Точечными линиями отмечены вероятности узнавания ($q = 1$, RG), вспоминания с подсказкой (пример для $q = 1/3$, CR) и свободного вспоминания ($q = 0$, FR).

Рис.1b. Распределение $D(P_{FR})$ частот, с которыми реализуется вероятность свободного вспоминания $P = P_{FR}$ при случайном выборе разных наборов из $N_d = 10$ разрываемых межнейронных связей для сети, характеристики которой приведены в подписи к рис.1a. Шкала по вертикали такая же, как на рис.1a. По горизонтали отложены значения $D(P_{FR})$ в процентах. Длина столбика на гистограмме соответствует вероятности реализации того значения $P_{FR}$, которое находится слева, напротив его основания на рис. 1a. Горизонтальная штриховая линия соединяет значение $P_{FR}$ для кривых 3,4 на рис.1a со значением (длина столбика гистограммы) вероятности их суммарной реализации (рис.1b). Справа от точечных линий лежат значения $D(P_{FR})$, которые >0.1%, >1%, >10% соответственно. Приведены также все значения $D(P_{FR})$, которые больше 10%.

Интересно отметить, что с вероятностью немногим менее 0.1% может реализоваться такая комбинация $N_d$ разорванных межнейронных связей, при которой вероятностные характеристики вспоминания информации, хранящейся в обученной сети, такие же как и для сети без повреждений (ей соответствует длина самого верхнего столбика на гистограмме на рис.1b, расположенного напротив самого правого кружка на кривой 1 рис.1a).

## 4. Заключение

В рамках количественной нейросетевой модели явления НКЯ [2,4] методом компьютерного моделирования продемонстрирована возможность существования та ких состояний в сетях с частью разорванных межнейронных связей. Таким образом, в [2,4] и в настоящей работе исследованы НКЯ, природа которых обусловлена:

1. Отсутствием в сети части входных нейронов вместе со всеми своими связями. При этом предполагается, что все межнейронные связи оставшихся нейронов сохранены полностью.
2. Разрывом в сети части ее межнейронных связей. При этом предполагается, что все нейроны в наличии и свои основные свойства сохраняют полностью.

Состояния НКЯ первого типа, связанные с отсутствием части нейронов входного слоя рабочей сети, преобладают. Они могут возникать в результате неполной (неадекватной) локализация нужной нейронной сети из состава словарного узла (см. раздел 2 этап 1), естественного забывания, болезни или травмы. Возникающие по всем этим причинам состояния НКЯ обладают одинаковыми характеристиками вспоминания. Однако затяжные, в конечном счете успешно разрешаемые состояния НКЯ (в качестве примера см. рассказ А.П. Чехова «Лошадиная фамилия» [3] и его нейросетевой анализ [2]) могут быть обусловлены только неадекватной локализацией сети, так как при предоставлении дополнительной подсказки рабочая сеть может быть локализована заново с исправлением имеющихся в ней дефектов [2]. Другие дефекты сети в процессе вспоминания не могут быть исправлены и поэтому могут быть ответственны только за обычные состояния НКЯ.

Состояния НКЯ второго типа, связанные с разрывом части межнейронных связей сети, возникают с вероятностью приблизительно в $10^3$ раз меньшей, чем НКЯ первого типа, однако это не означает, что факт их наличия имеет в только теоретическое значение. Вызывающие их дефекты могут быть обусловлены естественным забыванием, болезнью или травмой, сохраняются в процессе вспоминания и не могут быть причиной (см. выше) появления затяжных НКЯ. Не исключено, что при более полном численном исследовании сетей с разорванными межнейронными связями могут быть установлены еще менее вероятные состояния НКЯ.

## Литература

1. Гопыч П.М. (1999). Определение характеристик памяти. *Краткие сообщения ОИЯИ*, **4[96]-99**, 61-68.
2. Гопыч П.М. (2001). Трехэтапная количественная нейросетевая модель явления «на кончике языка». *Труды IX-й Международной конференции "Знание-диалог-решение" (KDS-2001), 19-22 июня 2001, С-Петербург, Россия*, 158-165. См. также: http://arXiv.org/abs/cs.CL/0107012.
3. Чехов А.П. (1885). Лошадиная фамилия. См. например: Сочинения в четырех томах. М., Правда, 1984, т.1, 188-192. См. также: *Russian Silhouettes: More Stories of Russian Life,* by Anton Tchekoff, translated from the Russian by Marian Fell. New York: Charles Scribner's Sons, October, 1915.
4. Gopych P.M. (2000). Quantitative Neural Network Model of the Tip-of-the-Tongue Phenomenon Based on Synthesized Memory-Psycholinguistic-Metacognitive Approach. *Proceedings of the 2nd International Conference 'Internet-Education-Science-2000', October 9-11, 2000, Vinnytsya, Ukraine*, 273. См. также: http://cogprints.soton.ac.uk/documents/disk0/00/00/10/33/cog00001033-00/IES_2000.txt.
5. *Koriat A. (2000). The feeling of knowing: some metatheoretical implications for consciousness and control. Consciousness and Cognition,* 9, 149-171.
6. *Levelt W.J.M. (1999). Models of word production. Trends in Cognitive Sciences,* **3**, 223-232.
7. Schwartz B.L. (1999). Sparking at the end of the tongue: The etiology of tip-of-the-tongue phenomenology. *Psychonomic Bulletin & Review,* **6**, 379-393.
8. Tulving E. (1989). Memory: Performance, knowledge, and experience. *European Journal of Cognitive Psychology,* **1**, 3-26.
9. Vigliocco, G. (2000) Language processing: The anatomy of meaning and syntax. *Current Biology,* **10**, R78-R80.